\documentclass[lettersize,journal]{IEEEtran}
\usepackage{amsmath,amsfonts}
\usepackage{algorithmic}
\usepackage{algorithm}
\usepackage{array}
\usepackage[caption=false,font=normalsize,labelfont=sf,textfont=sf]{subfig}
\usepackage{textcomp}
\usepackage{stfloats}
\usepackage{url}
\usepackage{verbatim}
\usepackage{graphicx}
\usepackage{cite}
\usepackage{multirow}
\usepackage{amsmath,amsfonts}
\usepackage{algorithmic}
\usepackage{algorithm}
\usepackage{array}
\usepackage{booktabs}
\usepackage[caption=false,font=normalsize,labelfont=sf,textfont=sf]{subfig}
\usepackage{textcomp}
\usepackage{stfloats}
\usepackage{url}
\usepackage{hyperref}
\usepackage{verbatim}
\usepackage{graphicx}
\usepackage{multirow}
\usepackage{hyperref}
\hypersetup{
    colorlinks=true,
    linkcolor=blue,
    filecolor=blue,      
    urlcolor=blue,
    citecolor=blue,
}

\begin{document}

\title{Learning A Coarse-to-Fine Diffusion Transformer for Image Restoration}

\author{Liyan Wang, Qinyu Yang, Cong Wang, Wei Wang, 
\\
Jinshan Pan,~\IEEEmembership{Member,~IEEE,} and Zhixun Su,~\IEEEmembership{Member,~IEEE}
\thanks{Liyan Wang, Qinyu Yang, and Zhixun Su are with School of Mathematical Sciences, Dalian University of Technology, Dalian, China; Zhixun Su is also with the Key Laboratory for Computational Mathematics and Data Intelligence of Liaoning Province and the Liaoning Center for Applied Mathematics, China (E-mail: zxsu@dlut.edu.cn).}

\thanks{Cong Wang is with the Department of Computing, The Hong Kong Polytechnic University, Hong Kong, China (E-mail: supercong94@gmail.com).}

\thanks{Wei Wang is with Dalian University of Technology, Dalian, China (E-mail: weiwei.transfer@gmail.com).}

\thanks{Jinshan Pan is with the School of Computer Science and Engineering, Nanjing University of Science and Technology, Nanjing, China (E-mail: sdluran@gmail.com).}}

\markboth{Journal of \LaTeX\ Class Files,~Vol.~14, No.~8, August~2021}%
{Shell \MakeLowercase{\textit{et al.}}: A Sample Article Using IEEEtran.cls for IEEE Journals}


\maketitle

\begin{abstract}
Recent years have witnessed the remarkable performance of diffusion models in various vision tasks. However, for image restoration that aims to recover clear images with sharper details from given degraded observations, diffusion-based methods may fail to recover promising results due to inaccurate noise estimation. Moreover, simple constraining noises cannot effectively learn complex degradation information, which subsequently hinders the model capacity. To solve the above problems, we propose a coarse-to-fine diffusion Transformer (C2F-DFT) for image restoration. Specifically, our C2F-DFT contains diffusion self-attention (DFSA) and diffusion feed-forward network (DFN) within a new coarse-to-fine training scheme. The DFSA and DFN respectively capture the long-range diffusion dependencies and learn hierarchy diffusion representation to facilitate better restoration. In the coarse training stage, our C2F-DFT estimates noises and then generates the final clean image by a sampling algorithm. To further improve the restoration quality, we propose a simple yet effective fine training scheme. It first exploits the coarse-trained diffusion model with fixed steps to generate restoration results, which then would be constrained with corresponding ground-truth ones to optimize the models to remedy the unsatisfactory results affected by inaccurate noise estimation. Extensive experiments show that C2F-DFT significantly outperforms diffusion-based restoration method IR-SDE and achieves competitive performance compared with Transformer-based state-of-the-art methods on $3$ tasks, including image deraining, image deblurring, and real image denoising. Code is available at \url{https://github.com/wlydlut/C2F-DFT}. 
\end{abstract}

\begin{IEEEkeywords}
Image Restoration, Diffusion Model, Transformer, Coarse-to-Fine Training.
\end{IEEEkeywords}

\section{Introduction}
\IEEEPARstart{I}{mage} restoration aims to recover clean images from low-quality ones affected by various degradation factors, such as rain, noise, blur, and more. 
The need to restore high-quality images for post-processing vision applications has driven significant interest in image restoration research.
Early conventional approaches in this field often rely on designing various statistical observations to properly formulate the problems~\cite{BM3D,pan2016blind,pan_pami_l0,Hu_2014_CVPR,Huang_2015_CVPR,li2016rain}. 
While these methods can achieve partial image recovery, they typically involve solving optimization algorithms that are challenging due to the non-convexity and non-smooth nature of the problems.
The advent of Convolutional Neural Networks (CNNs), capable of learning implicit priors from large-scale data, has led to the development of recent image restoration methods~\cite{RenZHZM19,deblurgan_cvpr18,dcsfn-wang-mm20,jdnet-wang-mm20,0002LZCC21,9847265}. 
However, CNNs have limitations due to their local receptive field and translation equivariance properties, which restrict their ability to model long-range pixel dependencies effectively.
In contrast, Transformers~\cite{vision_transformer,khan2021transformers}, with their attention mechanism that calculates responses at a given pixel through a weighted sum of all other positions, have emerged as a promising alternative.
This attention design allows Transformers to capture long-range dependencies, making them powerful for image restoration tasks. 
Consequently, Transformer-based approaches have been successfully applied to image restoration tasks and achieved impressive performance~\cite{wang2021uformer,Zamir2021Restormer}.

\begin{figure}[t]
\centering 
\includegraphics[scale=0.25]{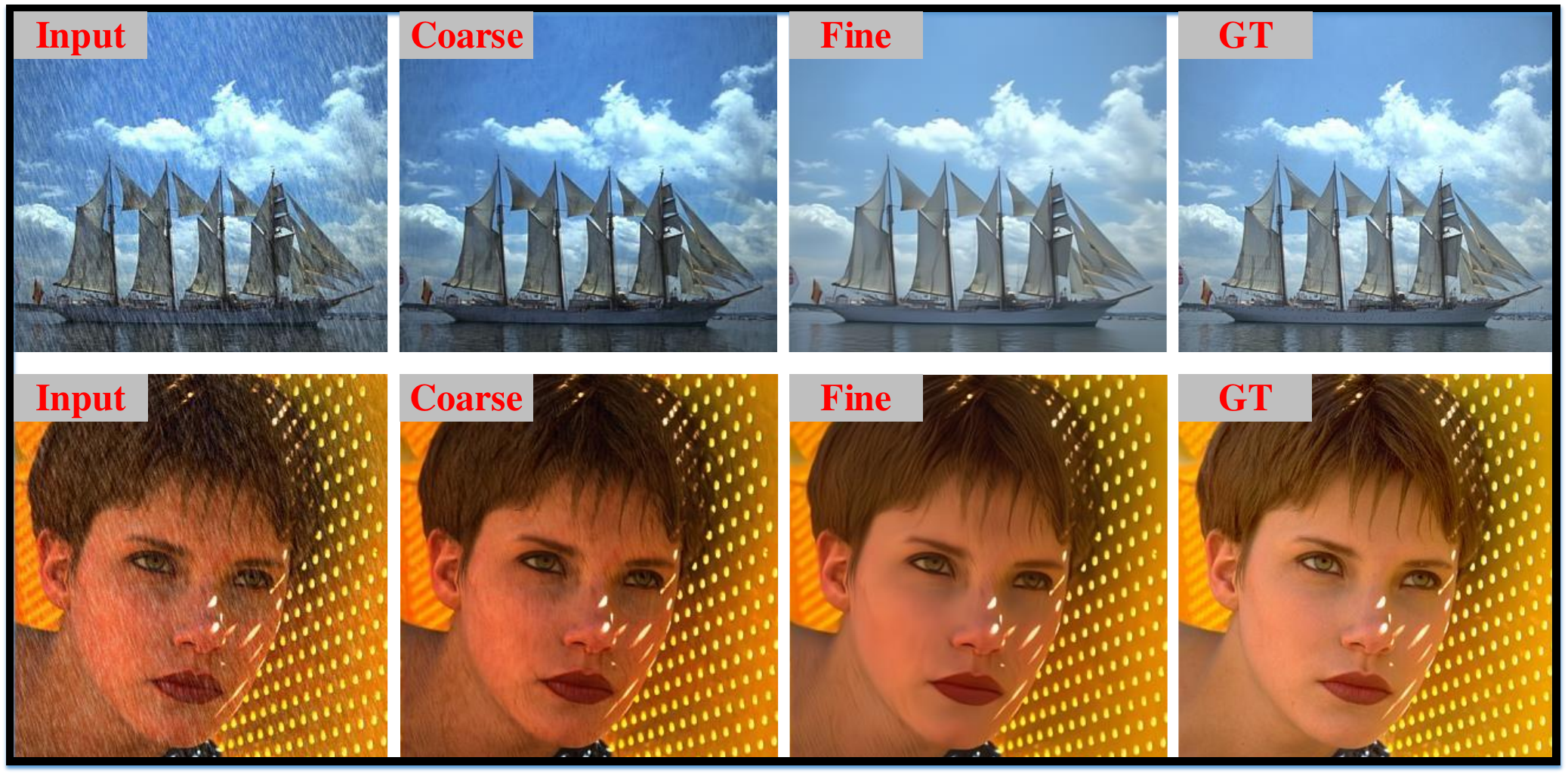}
\caption{
\underline{\textbf{Coarse training vs. fine training.}}
Unlike existing diffusion-based restoration models, e.g., \cite{valanarasu2022transweather}, which only constrain noises and obtain final restored results via a sampling algorithm (denoted as coarse training) that may fail to restore promising results due to inaccurate noise estimation, we propose the fine training to effectively improve restoration quality.}
\label{fig:teaser}
\end{figure}

Recently, diffusion model~\cite{HoJA20} has garnered considerable attention for its powerful generative capability and remarkable performance across various vision tasks, such as image generation~\cite{RombachBLEO22,0011SKKEP21}, inpainting~\cite{Xie_2023_CVPR_diffusion_inpainting}, detection~\cite{chen2022diffusiondet}, medical image segmentation~\cite{rahman2023ambiguous}, and also image restoration~\cite{cui2022semi,ChoiKJGY21,SahariaHCSFN23,WhangDTSDM22,OzdenizciL23,irsde}.
Unlike CNN and Transformer-based restoration methods that directly estimate final clear images from deep models, diffusion-based restoration models gradually recover clean images from noisy images generated in the forward diffusion process.
However, we note these methods are often trained by constraining noises and then directly obtain final clean images by a sampling algorithm. 
This training approach limits the model capacity as the simple estimation of noises may introduce inaccuracies that subsequently affect the sampling restoration quality, as shown in Fig.~\ref{fig:teaser}.

To solve the above problems, we propose the C2F-DFT, a diffusion Transformer (DFT) with a new coarse-to-fine (C2F) training scheme for image restoration.
Specifically, the C2F-DFT is built with diffusion Transformer blocks that contain diffusion self-attention (DFSA) and diffusion feed-forward network (DFN), where the time step is embedded into DFSA and DFN, to respectively capture the long-range diffusion dependencies and learn hierarchy diffusion features to facilitate better restoration.
To remedy the inaccuracy estimation of noise for restoration quality at the sampling process, we propose the coarse-to-fine training scheme.
The coarse-to-fine training scheme contains (a) coarse training and (b) fine training.
The coarse training is to train the diffusion Transformer by constraining noises, which would be exploited to obtain final restored images by a sampling algorithm.
To further improve the restoration quality, we propose a simple yet effective fine training to further optimize the DFT by constraining the sampled restored images with corresponding ground truth ones instead of noises to avoid generating unsatisfactory results due to inaccurate noise estimation.
With such designs, our fine training scheme can significantly improve the restoration quality compared with directly constraining noises in the coarse training, as shown in Fig.~\ref{fig:teaser}.

The main contributions of this paper can be summarized as follows:
\begin{itemize}
    \item We propose a diffusion Transformer for image restoration, which nicely embeds the diffusion into the Transformers, enabling it to not only model long-range dependencies but also fully exploit the generative ability of the diffusion model to facilitate better image restoration.
    
    \item We propose the coarse-to-fine training scheme to improve the restoration quality affected by the inaccurate noise estimation of diffusion models in coarse training, enabling to further scale up the model capacity in the fine training stage to facilitate better recovery.
     
    \item Extensive experiments show our C2F-DFT significantly outperforms recent diffusion-based method IR-SDE and achieves competitive performance compared with Transformer-based state-of-the-art methods on $3$ tasks, including image deraining, image deblurring, and real image denoising.
\end{itemize}

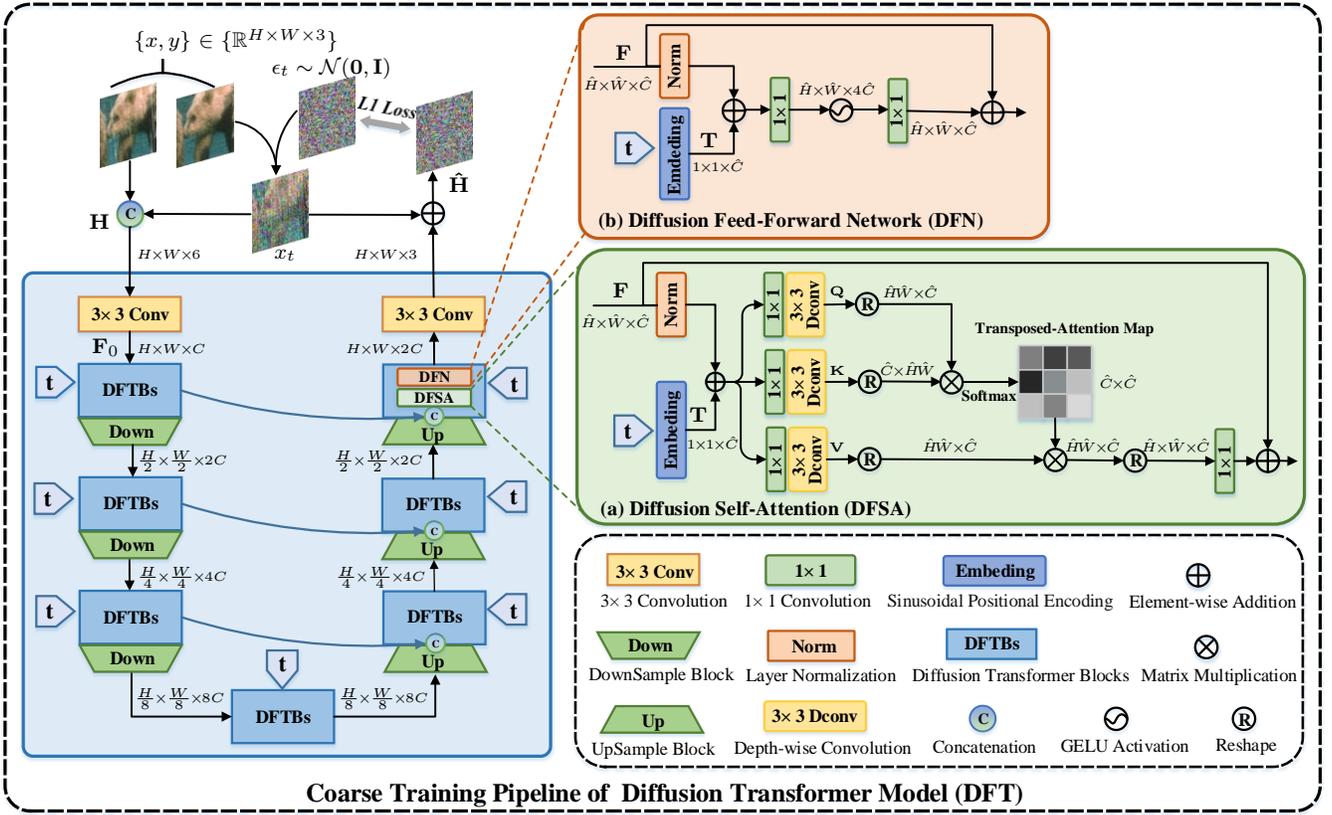
\begin{figure*}[t]
\centering
\includegraphics[width=2.0\columnwidth]{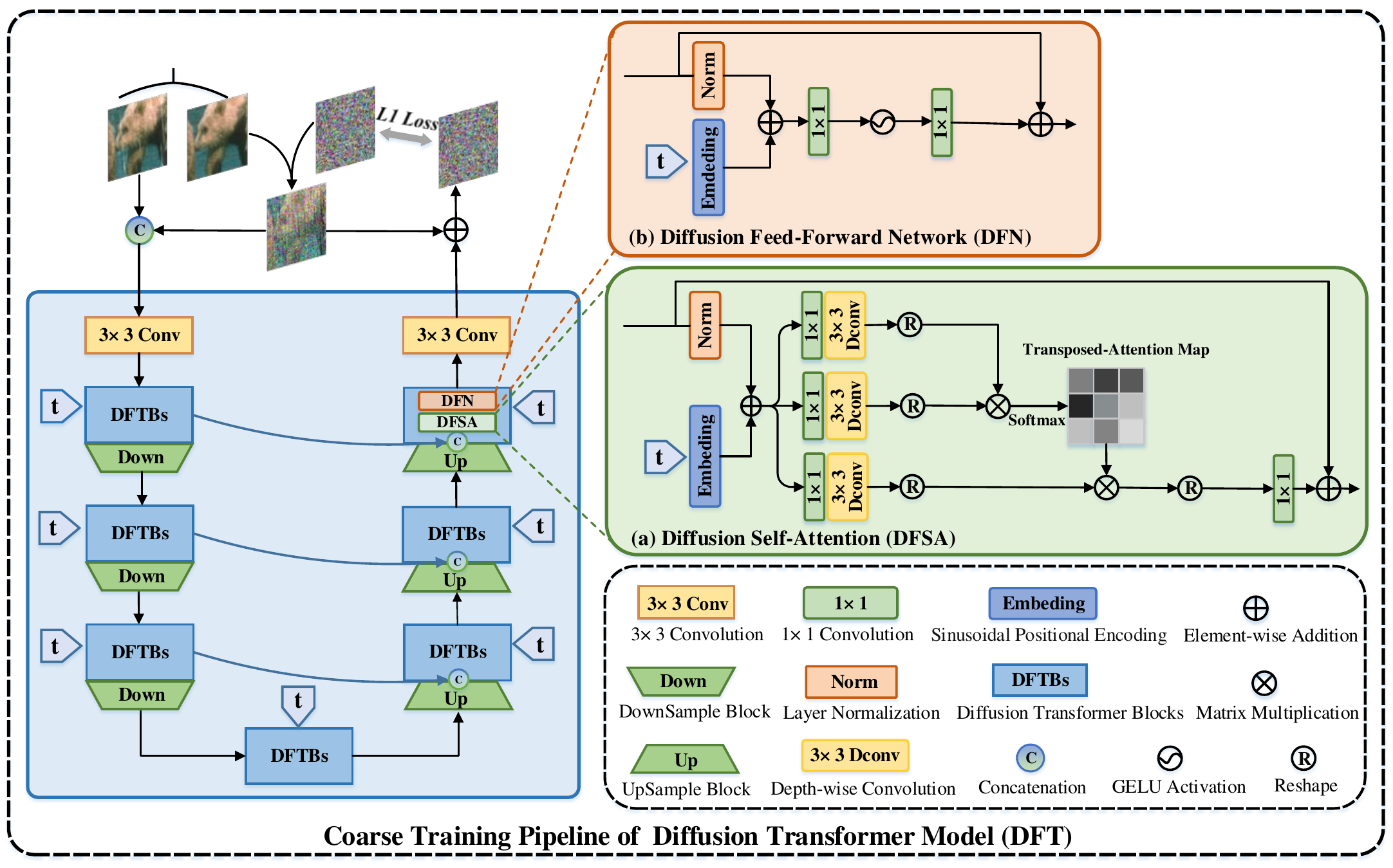}
\put(-452,294)
{\footnotesize{$\{x,y\}$~$\in$~$\{\mathbb{R}^{H\times W \times 3}\}$}}
\put(-400,284){\footnotesize{$\epsilon_t \sim \mathcal{N}(\mathbf{0},\mathbf{I}) $}}
\put(-399,214){\footnotesize{$x_{t}$}}
\put(-469,225){\footnotesize{$\mathbf{H}$}}
\put(-333,239){\footnotesize{$\mathbf{\hat{H}}$}}
\put(-451,214){\tiny{${H\!\!\times \!\!W \!\!\times \!\!6}$}}
\put(-369,214){\tiny{${H\!\!\times \!\!W \!\!\times \!\!3}$}}
\put(-468,178){\footnotesize{$\mathbf{F}_{0}$}}
\put(-451,178){\tiny{${H\!\!\times \!\!W \!\!\times \!\!C}$}}
\put(-451,136){\tiny{${\frac{H}{2}\!\!\times \!\!\frac{W}{2} \!\!\times \!\!2C}$}}
\put(-451,92){\tiny{${\frac{H}{4}\!\!\times \!\!\frac{W}{4} \!\!\times \!\!4C}$}}
\put(-452,46){\tiny{${\frac{H}{8}\!\!\times \!\!\frac{W}{8} \!\!\times \!\!8C}$}}
\put(-375,46){\tiny{${\frac{H}{8}\!\!\times \!\!\frac{W}{8} \!\!\times \!\!8C}$}}
\put(-376,92){\tiny{${\frac{H}{4}\!\!\times \!\!\frac{W}{4} \!\!\times \!\!4C}$}}
\put(-377,136){\tiny{${\frac{H}{2}\!\!\times \!\!\frac{W}{2} \!\!\times \!\!2C}$}}
\put(-372,178){\tiny{${H\!\!\times \!\!W\!\!\times\!\!2C}$}}
\put(-270,289){\footnotesize{$\mathbf{F}$}}
\put(-282,278){\tiny{${\hat{H}\!\!\times\!\!\hat{W}\!\!\times\!\!\hat{C}}$}}
\put(-238,256){\footnotesize{$\mathbf{T}$}}
\put(-201,276)
{\tiny{${\hat{H}\!\!\times\!\!\hat{W}\!\!\times\!\!4\hat{C}}$}}
\put(-241,247)
{\tiny{${1\!\!\times\!\!1\!\!\times\!\!\hat{C}}$}}
\put(-159,261){\tiny{${\hat{H}\!\!\times\!\!\hat{W}\!\!\times\!\!\hat{C}}$}}
\put(-271,199){\footnotesize{$\mathbf{F}$}}
\put(-283,188){\tiny{${\hat{H}\!\!\times\!\!\hat{W}\!\!\times\!\!\hat{C}}$}}
\put(-242,152){\footnotesize{$\mathbf{T}$}}
\put(-243,142){\tiny{${1\!\!\times\!\!1\!\!\times\!\!\hat{C}}$}}
\put(-189,200){\tiny{$\mathbf{Q}$}}
\put(-189,170)
{\tiny{$\mathbf{K}$}}
\put(-189,141)
{\tiny{$\mathbf{V}$}}
\put(-169,199){\tiny{${\hat{H}\!\hat{W}\!\!\times\!\!\hat{C}}$}}
\put(-170,170)
{\tiny{${\hat{C}\!\!\times\!\!\hat{H}\!\hat{W}}$}}
\put(-154,141){\tiny{${\hat{H}\!\hat{W}\!\!\times\!\!\hat{C}}$}}
\put(-100,141){\tiny{${\hat{H}\!\hat{W}\!\!\times\!\!\hat{C}}$}}
\put(-71,141){\tiny{${\hat{H}\!\!\times\!\!\hat{W}\!\!\times\!\!\hat{C}}$}}
\put(-87,165){\tiny{${\hat{C}\!\!\times\!\!\hat{C}}$}}
\caption{\underline{\textbf{Coarse Training Pipeline of Diffusion Transformer Model (DFT)}}.
We first add Gaussian noise to the clean image to obtain the noisy image according to the forward process of diffusion model~\cite{HoJA20} and concatenate it with the degraded images as the input of DFT.
Then, the DFT is optimized by constraining the noise, and we can obtain the final clean images via a sampling algorithm.
(a) Diffusion Self-Attention (DFSA) and (b) Diffusion Forward-Network (DFN) together build the diffusion Transformer block (DFTB), which respectively capture the long-range diffusion dependencies and learn hierarchy diffusion features to facilitate better restoration.
}
\label{fig:overall}
\end{figure*}

\section{Related Work}
In this section, we briefly talk about the related works of image restoration methods and diffusion models.

\subsection{Image Restoration}
Image Restoration (IR) aims to restore a clean image from its degraded observation. Traditional image restoration methods are mostly based on hand-crafted prior knowledge, such as sparse coding~\cite{LuoXJ15}, self-similarity~\cite{BuadesCM05}, gradient prior~\cite{XuZJ13}, etc. While these methods can achieve attractive performance on synthesized data, they struggle to cope with real-world image restoration due to the  limitation of the robustness and generalization capability.
With the emergence of Convolutional Neural Networks~\cite{AlexNet,he2016deep}, CNN-based image restoration methods~\cite{RIDNet,kim2020aindnet,ZamirAKHK0020,ZamirAKHKYS20,9302656} have achieved remarkable progress due to the powerful implicit learning ability from large-scale data. 
Since pioneering work SRCNN~\cite{DongLHT16} for image super-resolution, a flurry of CNN-based models has been proposed to improve model representation ability. Zhang et al.~\cite{ZhangZCM017} design the DnCNN network for image noising, which focuses on learning noise images rather than directly predicting the denoised image. To mitigate the gap between synthetic and real noise, CBDNet~\cite{CBDNet} inverses the demosaicing and gamma correction steps in image signal processing and then synthesizes signal-dependent Poisson-Gaussian noise in raw space.
Li et al.~\cite{LiWLLZ18} introduce a convolutional and recurrent neural network-based way named RESCAN to make full use of contextual information for image deraining.
Lee et al.~\cite{10061365} propose a new method that uses a locally adaptive channel attention module for a spectral–spatial network to resolve the problem of single-image deblurring.
Up to now, some widely used networks and structures in computer vision have been applied, including ResNet~\cite{ZhangTKZF21} with skip connections, the UNet~\cite{ZhangLZZGT22}
based on the encoder-decoder, the Attention Module~\cite{ZhangLLZF19}
that pays attention to the information of interest, and the GAN~\cite{ESRGAN,10083157} is designed to be a game between the discriminator and generator. 
 Most of them belong to end-to-end single-level mapping, highlighting the design and utilization of the model. Subsequently, the multi-stage image restoration methods~\cite{Zamir_2021_CVPR_mprnet,RenZHZM19,10077436} are proposed, which achieved great progress by recovering clean images in a progressive manner by employing a subnetwork at each stage.
Recently, Transformer~\cite{vision_transformer} has been widely applied to image restoration due to learning the long dependencies between image patch sequences and capturing the global interaction information between contexts~\cite{liang2021swinir,wang2021uformer,Zamir2021Restormer}. 
We refer the readers to the excellent literature review on image restoration~\cite{anwar2019deep,li2019single,tian2020deep,cho2021rethinking_mimo,purohit2021spatially_spair}, which summarise the main designs in deep image restoration models.

\subsection{Diffusion Models}
The originator of the diffusion model is the denoise diffusion probability model (DDPM) \cite{HoJA20}, which consists of two Markov chains. One disturbs the data into a forward chain of noise and the other converts the noise back to the reverse chain of the data, and then uses variational inference to gradually generate samples consistent with the original data distribution after a limited time. Recently, the diffusion model has achieved impressive performance due to its powerful generation ability.
Inspired by conditional generation, conditional diffusion models have been widely used in image restoration, such as image super-resolution~\cite{ChungSY22,LiYCCFXLC22,irsde}, image repair~\cite{EsserRBO21,JingCBJ22}, etc. 
Saharia et al. \cite{SahariaHCSFN23} employ a conditional diffusion model to generate realistic high-resolution images. However, it has to train a fixed low- to high-resolution model, and its adaptability to other resolutions is not strong enough.
Qzan et al.~\cite{OzdenizciL23} propose a patch-based denoising diffusion model for image restoration, but it requires longer sampling times due to multiple overlapping fixed-resolution image patches being input into the network multiple times.
Luo et al.~\cite{irsde} present a mean-reverting SDE-based method, which gradually restores a given low-quality image by simulating the reverse-time SDE for multiple steps. However, the training schedule and sampling steps are the same leading to increasing the computational cost at test time.
To reduce the computing needs of the training diffusion model for high-resolution images, LDM~\cite{RombachBLEO22} encodes the image into the hidden space for diffusion, which may not be conducive to mining the pixel information of the image itself. 
Under the LDM framework, Peebles et al. \cite{Dit} successfully replace the UNet backbone with a Transformer, which has more effective scalability. 

Although these works have achieved better performance, these methods usually are constrained by noises, which limits the model capacity as the simple estimation of noises may introduce inaccuracies that subsequently affect the sampling restoration quality.
In this paper, we remedy this problem by building a newly proposed coarse-to-fine training scheme for better image restoration.
Moreover, leveraging fixed sampling steps to construct a progressive restoration process is less explored.

\section{Diffusion Model Preliminaries}
As one of the classic unconditional generation models, DDPM~\cite{HoJA20} provides basic theoretical support for subsequent unconditional diffusion and conditional diffusion. Specifically, the forward diffusion process follows the $T$ steps Markov chain and gradually adds Gaussian noise according to a variance schedule $\{\beta_{1},\beta_{2},\cdots,\beta_{T}\}$ to the clean sample $\mathbf{x}_{0}\sim q(\mathbf{x}_{0})$. When $T$ is large enough, $\mathbf{x}_{T}$ is close to pure Gaussian noise. The process is:
\begin{equation} \label{eq1}
    \begin{aligned}
         &q(\mathbf{x}_{t}|\mathbf{x}_{t-1})=\mathcal{N}(\mathbf{x}_{t};\sqrt{1-\beta_{t}}\mathbf{x}_{t-1},\beta_{t}\mathbf{I}), \\&q(\mathbf{x}_{1:T}|\mathbf{x}_{0})=\prod_{t=1}^{T}q(\mathbf{x}_{t}|\mathbf{x}_{t-1}),
    \end{aligned}
\end{equation}
where $t\sim{\{0,\cdots,T\}}$; $\mathbf{x}_{t}$ is the noise image at time setp $t$. 

Considering that the forward diffusion process admits sampling $x_t$ at an arbitrary timestep $t$ in closed form,
\begin{equation} \label{eq3}
    \begin{aligned}
      q(\mathbf{x}_{t}|\mathbf{x}_{0})=\mathcal{N}(\mathbf{x}_{t};\sqrt{\bar{\alpha_{t}}}\mathbf{x}_{0},(1-\bar{\alpha_{t}})\mathbf{I}),
    \end{aligned}
\end{equation}
where, $\alpha_{t}=1-\beta_{t}$, $\bar{\alpha_{t}}=\prod_{i=1}^{t}\alpha_{i}$. 
It should be noted that the training objective completely depends on Eq.~\eqref{eq3}. 
By using the reparameterization trick, we can sample some images $\mathbf{x}_{t}$ at any time step $t$: $\mathbf{x}_{t}=\sqrt{\bar{\alpha_{t}}}\mathbf{x}_{0}+\sqrt{1-\bar{\alpha_{t}}}\epsilon_{t}$, where $\epsilon_{t}\sim \mathcal{N}(\mathbf{0},\mathbf{I})$ has the same dimensionality as clean data $\mathbf{x}_{0}$  and latent variables $\mathbf{x}_{t}$.

In the reverse diffusion process, diffusion models are trained to learn the reverse process, that the joint distribution $p_{\theta}(\mathbf{x}_{0:T})$ defined by as a Markov chain with learned Gaussian transitions starting at $p(\mathbf{x}_{T})=\mathcal{N}(\mathbf{x}_{T};\mathbf{0},\mathbf{I})$:
\begin{equation} \label{eq4}
    \begin{aligned}
        &p_{\theta}(\mathbf{x}_{0:T})=p(\mathbf{x}_{T})\prod_{t=1}^{T}p_{\theta}(\mathbf{x}_{t-1}|\mathbf{x}_{t}),
        \\
        &p_{\theta}(\mathbf{x}_{t-1}|\mathbf{x}_{t})=\mathcal{N}(\mathbf{x}_{t-1};\mu_{\theta}(\mathbf{x}_{t},t),\Sigma_{\theta}(\mathbf{x}_{t},t)).
    \end{aligned}
\end{equation}

Here, \cite{HoJA20} integrates $\mu_{\theta}(\mathbf{x}_{t},t)$ reparameterize to noise prediction network $\epsilon_{\theta}(\mathbf{x}_{t},t)$ with trainable model parameter $\theta$:
\begin{equation} \label{eq6}
    \begin{aligned}
       \mu_{\theta}(\mathbf{x}_{t},t)=\frac{1}{\sqrt{\alpha_{t}}}(\mathbf{x}_{t}-\frac{1-\alpha_{t}}{\sqrt{1-\bar{\alpha_{t}}}}\epsilon_{\theta}(\mathbf{x}_{t},t)).
    \end{aligned}
\end{equation}

Because learning a diagonal variance $\boldsymbol{\Sigma}_\theta$ leads to unstable training and poorer quality, \cite{HoJA20} adopts a simplified objective:
\begin{equation} \label{eq7}
    \begin{aligned}
    \mathcal{L}_{t}(\theta)&=\mathbb{E}_{t\sim{[1,T]},\mathbf{x}_{0},\epsilon_{t}}[\parallel \epsilon_{t}-\epsilon_{\theta}(\mathbf{x}_{t},t)\parallel^{2}]\\&=\mathbb{E}_{t\sim {[1,T]},\mathbf{x}_{0},\epsilon_{t}}[\parallel \epsilon_{t}-\epsilon_{\theta}(\sqrt{\bar{\alpha_{t}}}\mathbf{x}_{0}+\sqrt{(1-\bar{\alpha_{t}}})\epsilon_{t},t)\parallel^{2}].
    \end{aligned}
\end{equation}

%
\section{Methodology}
Our goal aims to effectively train the diffusion Transformer for better restoration.
To that end, we propose a coarse-to-fine training pipeline to improve recovery quality.
\begin{figure*}[t]
 \centering 
 \includegraphics[width=1.5\columnwidth]{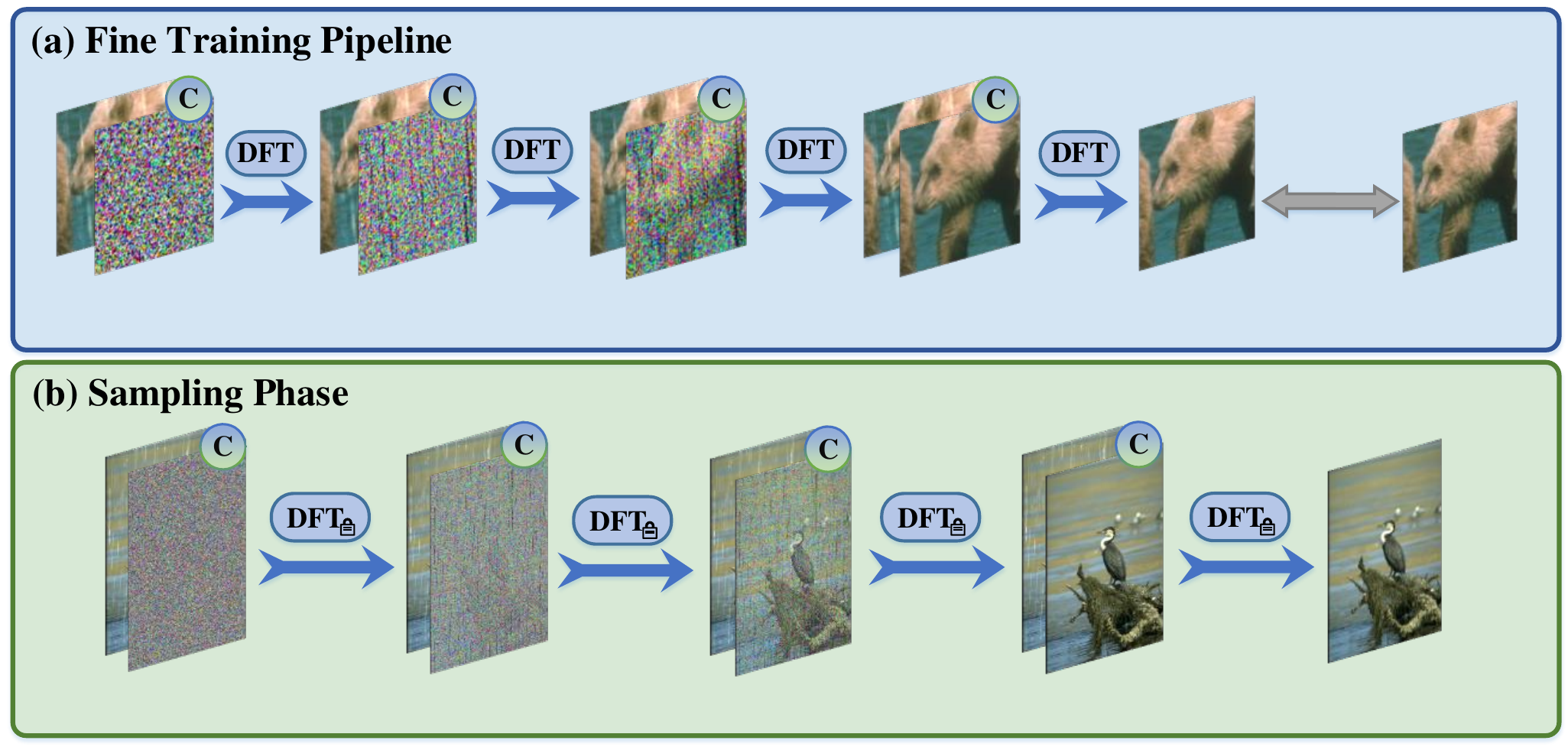}
\put(-358,105){{$(y,x_{t_4})$}}
\put(-293,105){{$(y,x_{t_3})$}}
\put(-229,105){{$(y,x_{t_2})$}}
\put(-164,105){{$(y,x_{t_1})$}}
\put(-92,105){{$x_{t_0}$}}
\put(-27,105){{$x$}}
\put(-67,140){{$\mathcal{L}_{rec}$}}
\put(-348,10){{$(y,x_{t_4})$}}
\put(-275,10){{$(y,x_{t_3})$}}
\put(-201,10){{$(y,x_{t_2})$}}
\put(-126,10){{$(y,x_{t_1})$}}
\put(-48,10){{$x_{t_0}$}}
  \caption{\underline{\textbf{(a) Fine Training Pipeline.}} 
  We first initialize the parameters of DFT with the well-trained parameters in the coarse training. 
  We then adopt the sampling algorithm to generate clean images, which would participate in optimizing DFT by constraining the sampling restoration results with corresponding ground truth ones.
  \underline{\textbf{(b) Sampling Phase.}} Our method can directly test the entire image with any size in a progressive restoration manner, greatly improving the restoration quality.
  }
  \label{fig: Illustration of the fine training pipeline. (b) Diagram of the sampling phase by a progressive restoration manner detailed in Algorithm 1.}
\end{figure*}
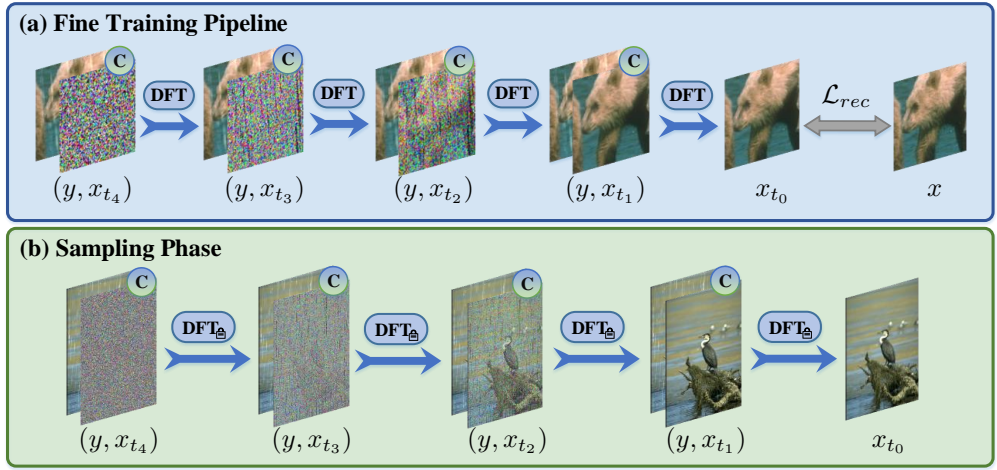
\subsection{Diffusion Transformer Model}
Fig.~\ref{fig:overall} shows the overall of our diffusion Transformer (DFT), which is a 4-level U-shaped structure with diffusion Transformer blocks (DFTBs). 
The DFTBs consist of diffusion self-attention (DFSA) and diffusion feed-forward network (DFN), as shown in Fig.~\ref{fig:overall}(a) and (b), respectively.

\subsubsection{Overall Pipeline}
Given the paired clean and degraded images $\{x,y\}$~$\in$~${\mathbb{R}^{H \times W \times 3}}$, where $H \times W$ denotes the spatial dimension, we first obtain the noise sample $x_{t}\in{\mathbb{R}^{H \times W \times 3}}$ by adding Gaussian noise $\epsilon_t \sim \mathcal{N}(\mathbf{0},\mathbf{I}) $ at time step $t$ on the clean image $x$ according to the forward process of diffusion model~\cite{HoJA20}, and concatenate $x_{t}$ with the degraded image $y$ at channel dimension to obtain $\mathbf{H}$~$\in$~$\mathbb{R}^{H \times W \times 6}$ as the input of DFT.
Next, $\mathbf{H}$ is encoded using a $3\times3$ convolution to obtain the embedding feature $\mathbf{F}_{0}$~$\in$~${\mathbb{R}^{H\times W\times C}}$, where $C$ means the number of channel. 
$\mathbf{F}_{0}$ is hierarchically encoded and decoded via DFTBs, while the time $t$ is encoded to the feature $\mathbf{T}$ which is further embedded into the DFTBs. 
We also utilize skip connections to connect the features at the same level in the encoder and decoder. 
Finally, we use a $3 \times 3$ convolution to obtain the residual image, which is added to $x_{t}$ to obtain the estimated noise $\mathbf{\hat{H}}$~$\in$~${\mathbb{R}^{H \times W \times 3}}$.
\subsubsection{Diffusion Self-Attention}
%
Our DFSA aims to model long-range diffusion dependencies.
Given the time step $t$ of the diffusion model, we use sinusoidal positional encoding~\cite{VaswaniSPUJGKP17} to encode $t$ into the vector embedding $\mathbf{T}$~$\in$~${\mathbb{R}^{1 \times 1 \times C}}$ (as the dimension of time $t$ is independent of image resolution, DFSA can process images with any sizes).
We then embed the $\mathbf{T}$ into given the input features $\mathbf{F}$ to conduct self-attention~\cite{Zamir2021Restormer}:
\begin{equation}\label{eq: Diffusion Self-Attention.}
    \begin{aligned}   
    &\textit{DFSA}\big(\mathbf{F},\mathbf{T}\big)= \mathcal{A}\big(\textbf{Q}, \textbf{K}, \textbf{V}\big)+\mathbf{F}, \\
&\textbf{Q}, \textbf{K}, \textbf{V} = \textit{Split}\Big(W_{d}W_{p}\big(LN(\mathbf{F})\big)+\mathbf{T}\Big),   
    \end{aligned}
\end{equation}
where $W_{p}$ and $W_{d}$ respectively denote $1$$\times$$1$ point-wise convolution and $3$$\times$$3$ depth-wise convolution;
$\mathcal{A}\big(\textbf{Q}, \textbf{K}, \textbf{V}\big) = \textbf{V}\cdot \textrm{Softmax$\big(\textbf{K} \cdot \textbf{Q}/\alpha \big)$}$; 
Here, $\alpha$ is a learnable scaling parameter to control the magnitude of the dot product of $\textbf{K}$ and $\textbf{Q}$ before applying the softmax function;
$\textit{Split}$ denotes the split operation;
$\textit{LN}(\cdot)$ means layer normalization~\cite{ba2016layer}.

\subsubsection{Diffusion Feed-Forward Network}
%
Our DFN is to learn hierarchy diffusion representation, which is achieved by exploiting the two $1 \times 1$ point-wise convolutions as well as the time step embedding $\mathbf{T}$ to further process the output features $\mathbf{F}$ of DFSA:
\begin{equation} \label{eq12}
    \begin{aligned}
   \textit{DFN}\big(\mathbf{F}, \mathbf{T}\big) = W_{p}\phi W_{p}\big(\textit{LN}(\mathbf{F})+\mathbf{T}\big)+\mathbf{F},   
    \end{aligned}
\end{equation}
where $\phi$ denotes the non-linear activation function GELU.

\subsection{Coarse-to-Fine Training Pipeline for Restoration}
Our coarse-to-fine training scheme contains (a) coarse training and (b) fine training.
The coarse training aims to train the diffusion Transformer by constraining noises, which would be exploited to obtain final restored images by a sampling algorithm, while the fine training further optimizes the diffusion Transformer by constraining the sampled clean images with fixed steps in the coarse training and corresponding ground truth ones to scale up the model capacity for better recovery.

\subsubsection{Coarse Training}
%
Our coarse training is similar to the existing conditional diffusion model, which is to estimate the noise $\epsilon_{t}$.
Hence, the loss function in coarse training is: 
\begin{equation} \label{eq: loss-Coarse Training}
\begin{aligned}
   \mathcal{L}_{t}(\theta)&=\mathbb{E}_{t\sim{[1,T]}, x_{0},\epsilon_{t}}\parallel\epsilon_{t}-\epsilon_{\theta}(x_{t},y,t)\parallel_{1},
\end{aligned}
\end{equation}
where the $\epsilon_{\theta}$ denotes the noise estimation network, i.e., our diffusion Transformer in this paper.
The estimated noise can be further exploited to produce final clean images via a sampling algorithm.

\begin{table*}[!b]
\caption{\underline{\textbf{Image deraining.}}
The CNN, TF, and DM respectively mean the CNN, Transformer, and diffusion model-based approaches.
$\text{IR-SDE}^{*}$ denotes we use the provided pre-trained models to test, which respectively are individually trained on Rain100H and Rain100L.
$\text{IR-SDE}$ means we use the publicly available codes to re-train, where we train IR-SDE on Rain13K for fair comparisons, then test on different test sets.
The \textbf{best} and \underline{second best} are marked in bold and underlined, respectively.
} \label{tab:Image deraining.}
\begin{center}
\setlength{\tabcolsep}{0.4mm}
\begin{tabular}{l|l|cccccccc||cc}
\toprule
\multicolumn{2}{l|}{\multirow{2}{*}{\textbf{Methods}}}
&\multicolumn{2}{c}{\textbf{Test100}\cite{ZhangSP20a}}  &\multicolumn{2}{c}{\textbf{Rain100H}~\cite{derain_jorder_yang}}   &\multicolumn{2}{c}{\textbf{Rain100L}~\cite{derain_jorder_yang}}  &\multicolumn{2}{c||}{\textbf{Test2800}~\cite{0004P18}} &\multicolumn{2}{c}{\textbf{Average}} 
\\
\multicolumn{2}{c|}{}
&~~~PSNR~$\uparrow$~ & ~~~SSIM~$\uparrow$~~
&~~~PSNR~$\uparrow$~& ~~~SSIM~$\uparrow$~~
&~~~PSNR~$\uparrow$~ & ~~~SSIM~$\uparrow$~~ 
&~~~PSNR~$\uparrow$~ & ~~~SSIM~$\uparrow$~~
&~~~PSNR~$\uparrow$~ & ~~~SSIM~$\uparrow$~~ \\
\toprule
\multirow{6}{*}{\textbf{CNN}} ~&~RESCAN\cite{LiWLLZ18} &25.00 & 0.835 &26.36 & 0.786 &29.80 & 0.881 &31.29 & 0.904 &28.11 & 0.851\\ 
&~PreNet\cite{RenZHZM19}  &24.81 & 0.851 &26.77 & 0.858 &32.44 & 0.950  &31.75 & 0.916 &28.94 & 0.893\\ 
&~MSPFN\cite{mspfn2020} &27.50 & 0.876 &28.66 & 0.860 &32.40 & 0.933 &32.82 & 0.930 &30.34 &  0.899\\
&~MPRNet\cite{Zamir_2021_CVPR_mprnet}  &30.27 & 0.897	&30.41 & 0.890	&36.40 & 0.965	&33.64 & 0.938 & 32.68 & 0.922\\
&~HINet\cite{0002LZCC21}  &30.29 & 0.906	&30.65& 0.894	&37.28 &0.970  &33.91 & \underline{0.941} & 33.03 & 0.927 \\
&~SPAIR\cite{purohit2021spatially_spair}  &30.35 & 0.909	&30.95 & 0.892	&36.93 & 0.969	&33.34 & 0.936 &32.89 &  0.926\\
\toprule
\multirow{1}{*}{\textbf{TF}} ~&~Restormer\cite{Zamir2021Restormer}  &\textbf{32.00} & \textbf{0.923}	&\underline{31.46} & \underline{0.904}	
&\underline{38.99} & \underline{0.978}	&\textbf{34.18}& \textbf{0.944} &\textbf{34.15} & \underline{0.937}\\
\toprule
\multirow{3}{*}{\textbf{DM}} ~&~$\text{IR-SDE}^{*}$\cite{irsde}  &	-  & -  &$31.65^{*}$  &  $0.904^{*}$	&$38.30^{*}$  &$0.980^{*}$	&    -   &    - &    -   &    -   \\
~&~$\text{IR-SDE}$\cite{irsde} &	 26.74  & 0.834  &20.79  &  0.699	&30.83  &0.912	&    30.42  &  0.891  & 27.20   &  0.834   \\
& ~\textbf{C2F-DFT (Ours)}~~~ &\underline{31.38}& \underline{0.921}	&\textbf{31.62} &  \textbf{0.909} &\textbf{39.03} & \textbf{0.980}	&\underline{34.03} & \textbf{0.944} &\underline{34.01} & \textbf{0.938}\\
\toprule
\end{tabular}
\end{center}
\vspace{-3mm}
\end{table*}
\subsubsection{Fine Training}
After completing the coarse training, we observe that 4-step sampling reaches the best restoration quality in terms of PSNR/SSIM, as shown in Tab.~\ref{tab:Effect on different sampling steps tested on the entire image.}. 
However, as the coarse training primarily focuses on constraining noises, inaccurate noise estimation may significantly impact restoration quality (see Fig.~\ref{fig:teaser}, Fig.~\ref{fig:Coarse vs. Fine training visual results on image deraining.} and Fig.~\ref{fig:Coarse vs. Fine training visual results on image deblurring.}).
To address this problem to achieve better restoration, we propose a simple yet effective fine training scheme to further optimize the model by constraining the sampled restoration results with fixed sampling steps instead of noises, as shown in Fig.~\ref{fig: Illustration of the fine training pipeline. (b) Diagram of the sampling phase by a progressive restoration manner detailed in Algorithm 1.}(a). 
The fine training stage has the same data processing as the coarse training but differs from the constraint objects.
Specifically, we first initialize the parameters of DFT with the well-trained parameters from the coarse training. 
We then incorporate a sampling algorithm with $4$-step to generate restoration results.
Last, we optimize DFT by constraining the generated sampling restoration images with the corresponding ground-truth ones using $L_1$ loss ($\mathcal{L}_{1}$) and SSIM loss ($\mathcal{L}_{ssim}$)~\cite{ZhaoGFK17} instead of constraining noise to remedy the unsatisfactory results in coarse training: 
\begin{equation} \label{eq: loss fine training}
    \begin{aligned}
     \mathcal{L}_{rec} = \lambda \mathcal{L}_{ssim}(x_{t_0},x)+(1-\lambda)\mathcal{L}_{1}(x_{t_0},x),
    \end{aligned}
\end{equation}
where $x_{t_{0}}$ denotes the generated sampled restoration results;
$\mathcal{L}_{ssim}(x_{t_0},x)$=$1-\text{SSIM}(x_{t_0},x)$; $\lambda$ is a weight.

We will demonstrate that the fine training phase has the capacity to surpass the performance limitations set by the diffusion model during the coarse training stage. This breakthrough not only paves the way for enhancing model capacity but also holds the potential to advance restoration tasks in future diffusion-based image restoration models.

\subsection{Sampling Algorithm}
During the sampling stage, we employ the implicit sampling strategy~\cite{SongME21} to expedite our sampling process. The procedure for sampling within our C2F-DFT is outlined in Alg.~\ref{alg:algorithm}. 
As we use the fine training scheme to train the DFT by constraining the sampled restoration results with corresponding ground truth ones instead of constraining noises, this advancement significantly enhances the quality of the samples.
Fig.~\ref{fig: Illustration of the fine training pipeline. (b) Diagram of the sampling phase by a progressive restoration manner detailed in Algorithm 1.}(b) shows our sampling outline, which progressively recovers clear images by the trained DFT. 
\begin{algorithm}[h]
\caption{The sampling process of our C2F-DFT}
\label{alg:algorithm}
\textbf{Input}: Degraded image $y$, diffusion Transformer $\epsilon_{\theta}(x_{t},y,t)$, time steps $T$, implicit sampling steps $S$.\\
\textbf{Output}: Restored image $x_{t_0}$.
\begin{algorithmic}[1] 
\STATE $x_{t_{s}}\sim{\mathcal{N}(0,\mathbf{I}})$
\FOR {$j=S, \cdots, 1$} 
\STATE $t_{j}= (j-1)\cdot T/(S-1) +1$
\STATE {$t_{j-1}=(j-2)\cdot
    T/(S-1) + 1$} \textbf{if} $j >1$ \textbf{else} $0$
\STATE $\hat{\epsilon}_{t_{j}}=\epsilon_{\theta}(x_{t_{j}},y,t_{j})$ 
\STATE $x_{t_{j-1}} = \sqrt{\bar{\alpha}_{t_{j-1}}}(\frac{x_{t_{j}}-\sqrt{1-\bar{\alpha}_{t_{j}}} \cdot \hat{\epsilon}_{t_{j}}}{\sqrt{\bar{\alpha}_{t_{j}}}})+\sqrt{1-\bar{\alpha}_{t_{j-1}}}\cdot \hat{\epsilon}_{t_{j}}$ 
\ENDFOR
\STATE \textbf{return} $x_{t_0}$
\end{algorithmic}
\end{algorithm}
\subsection{Patch-Cycle Diffusion Learning Strategy}

In contrast to existing diffusion models that rely on fixed patches for learning the diffusion process, we introduce a patch-cycle diffusion learning strategy, which enables diffusion models to capture more contextual information for better restoration. 
Specifically, we extract $p \times p$ patches from clean-degraded image pairs, denoted as $\{x^{p},y^{p}\}$, where $p$ is selected from $\{32, 64, 128\}$ in our experiments.
During the training phase, we cyclically input $\{x^{p},y^{p}\}_{\{p=32, 64, 128\}}$ into the C2F-DFT every $N$ iterations and continue this training strategy until completion. 
To manage training costs, we decrease the batch size as $p$ increases.

\begin{figure*}[t]
\centering
\includegraphics[scale=0.545]{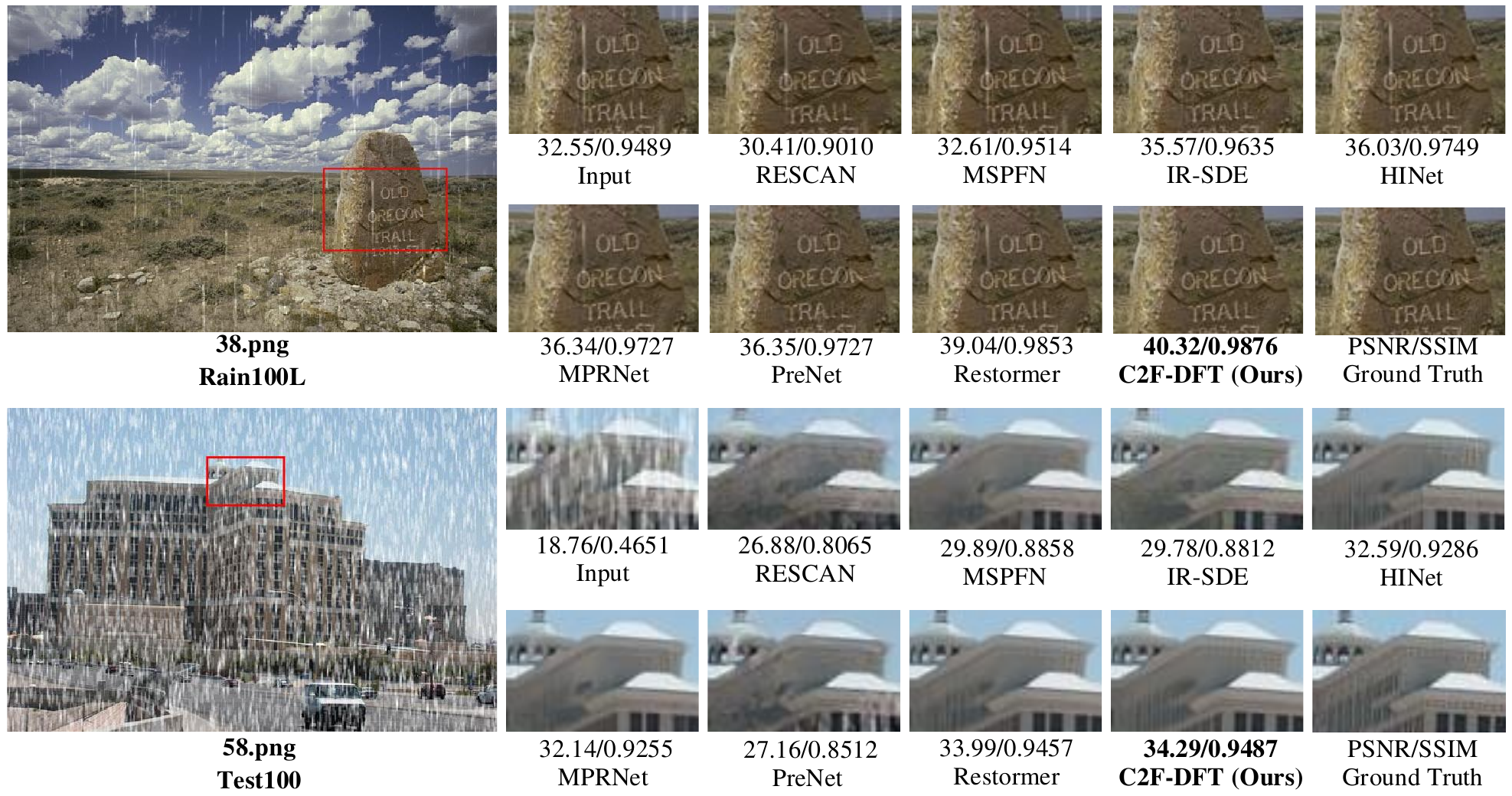}
\vspace{-5mm}
\caption{
\underline{\textbf{Image deraining on Rain100L~\cite{derain_jorder_yang} and Test100~\cite{ZhangSP20a}}}.
C2F-DFT is able to generate much cleaner results.}
\label{fig:Image deraining on Rain100L}
\end{figure*}

\begin{figure*}[t]
\centering
\includegraphics[scale=0.512]{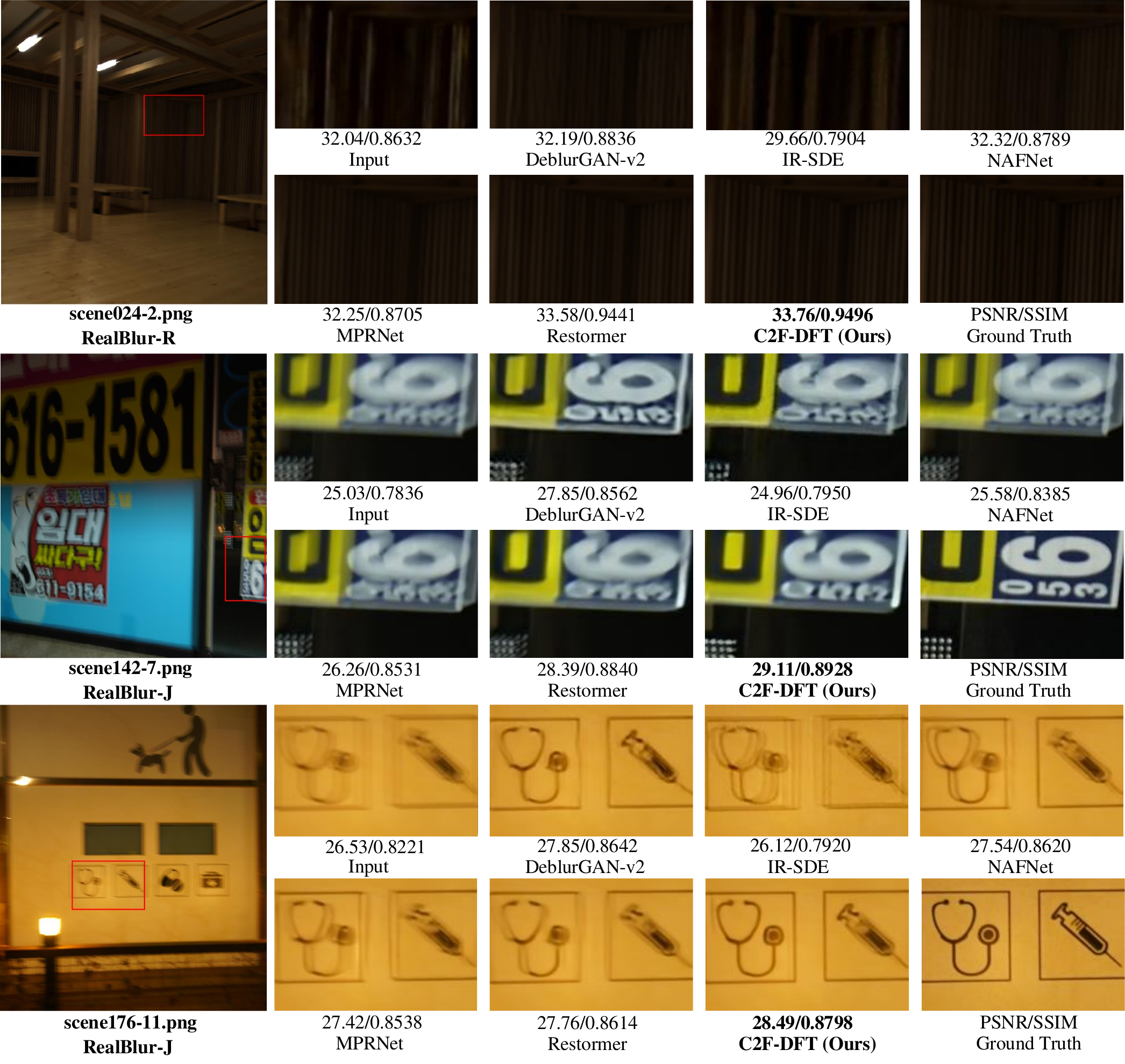} 
\vspace{-5mm}
\caption{\underline{\textbf{Image deblurring on RealBlur-R~\cite{rim_2020_realblur} and RealBlur-J~\cite{rim_2020_realblur}}}. 
C2F-DFT is able to generate results with sharper structures.}
\label{fig:Image deblurring on RealBlur-R}
\end{figure*}
\begin{table*}[!b]
\begin{center}
\caption{\underline{\textbf{Image deblurring.}}
Our method is trained on GoPro~\cite{gopro2017} and then directly applies to RealBlur-R~\cite{rim_2020_realblur} and RealBlur-J~\cite{rim_2020_realblur}.
} \label{tab:Image deblurring}
\setlength{\tabcolsep}{5.7mm}
\begin{tabular}{l|l|cc||cccc}
\toprule
\multicolumn{2}{l|}{\multirow{2}{*}{\textbf{Methods}}}&\multicolumn{2}{c||}{\textbf{GoPro}~\cite{gopro2017}} &\multicolumn{2}{c}{\textbf{RealBlur-R}~\cite{rim_2020_realblur}}&\multicolumn{2}{c}{\textbf{RealBlur-J}~\cite{rim_2020_realblur}}\\
\multicolumn{2}{c|}{} &PSNR~$\uparrow$ & SSIM~$\uparrow$
&PSNR~$\uparrow$ & SSIM~$\uparrow$
&PSNR~$\uparrow$ & SSIM~$\uparrow$ \\
\toprule
\multirow{9}{*}{\textbf{CNN}}&DeblurGAN\cite{deblurgan_cvpr18} &28.70&0.858 
&33.79&0.903&27.97&0.834\\
&DeblurGAN-v2\cite{deblurv2}  &29.55&0.934
&35.26&0.944  &28.70&0.866 \\ 
&SRN\cite{tao2018scale}  &30.26&0.934
&35.66&0.947 &28.56&0.867\\
&DBGAN\cite{zhang2020dbgan}  &31.10&0.942
&33.78&0.909 &24.93&0.745\\
&MT-RNN\cite{mtrnn2020}  &31.15&0.945
&35.79&0.951  &28.44&0.862\\
&DMPHN\cite{dmphn2019}  &31.20&0.940
&35.70&0.948 &28.42&0.860\\
&SPAIR\cite{purohit2021spatially_spair}  &32.06&0.953 &- & -
&28.81&0.875\\
&MIMO-UNet++\cite{cho2021rethinking_mimo}~~&32.45&0.957
&35.54&0.947 &27.63&0.837\\
&MPRNet\cite{Zamir_2021_CVPR_mprnet}  &32.66& 0.959
&35.99&0.952 &28.70&0.873\\
&NAFNet\cite{chen2022simple_naf}  &\textbf{33.71}& \textbf{0.967}
&35.97&0.951 &28.31& 0.856\\
\toprule
\multirow{1}{*}{\textbf{TF}}&Restormer\cite{Zamir2021Restormer}  &\underline{32.92}&\underline{0.961}
&\underline{36.19}&\underline{0.957} &\textbf{28.96}&\textbf{0.879}\\
\toprule
\multirow{2}{*}{\textbf{DM}}&IR-SDE~\cite{irsde}   &30.70&0.901 &33.96	&0.918 &24.21&0.729   \\
&\textbf{C2F-DFT (Ours)}&31.96&0.928  &\textbf{36.34}&\textbf{0.957} &\underline{28.90}&\underline{0.876} \\
\toprule
\end{tabular}
\end{center}
\end{table*}

\section{Experiments}
We evaluate C2F-DFT on $3$ popular image restoration tasks: \textbf{(a)} image deraining, \textbf{(b)} image deblurring, and \textbf{(c)} real image denoising. 
%
All of our models are trained on two NVIDIA RTX A6000 GPUs. Next, we present more implementation details and experimental results for each task. Note that the results of the comparison methods are obtained by using the official codes and the pre-trained model for fair comparisons.

\subsection{Implementation Details}

\subsubsection{Network Settings}
From level-1 to level-4, the number of DFTBs is [4, 6, 6, 8], attention heads in DFSA are [1, 2, 4, 8], and the number of channels is [48, 96, 192, 384]. 

\subsubsection{Diffusion Settings}
In diffusion training, we set the forward process variances to constants increasing linearly from $\beta_{1}=1e^{-4}$ to $\beta_{2}=2e^{-2}$, and the total time step $T$ is set to $1000$. 

\subsubsection{Coarse Training}
We train our deraining model using AdamW optimizer ($\beta_{1}$=$0.9$, $\beta_{2}$=$0.999$) with $270$K iterations.
We set the initial learning rate as $3e^{-4}$ which is gradually reduced to $1e^{-5}$ with the cosine annealing~\cite{LoshchilovH17}.
The (patch, batch) are cyclically updated by $[(32^2, 360), (64^2, 96), (128^2, 24)]$ at every $10$K iterations.  The deblurring model and denoising model are trained with the (patch size, batch size) cyclically updated by $[(64^2, 96), (128^2, 24), (256^2, 6)]$.

\subsubsection{Fine Training}
%
After finishing the coarse training, we further train our deraining model with $90$K iterations with the initial learning rate $1e^{-5}$ gradually reduced to $1e^{-7}$ for fine image restoration. 
The (patch, batch) are cyclically updated by $[(32^2, 96), (64^2, 24), (128^2, 6)]$ at every $5$K iterations. The deblurring model and denoising model are further trained with the (patch size, batch size) cyclically updated by $[(32^2, 24), (64^2, 6), (128^2, 1)]$.
We empirically set $\lambda = 0.84$ in Eq.~\eqref{eq: loss fine training}. 

\begin{table*}[t]
\begin{center}
\caption{\underline{\textbf{Real image denoising.}}
Our method is trained on the SIDD training set~\cite{sidd} and then is directly applied to DND~\cite{PlotzR17}.
} \label{tab:Real image denoising}
\setlength{\tabcolsep}{7.3mm}
\begin{tabular}{l|l|cc||cc}
\toprule
\multicolumn{2}{l|}{\multirow{2}{*}{\textbf{Methods}}}&  \multicolumn{2}{c||}{\textbf{SIDD}~\cite{sidd}}  &  \multicolumn{2}{c}{\textbf{DND}~\cite{PlotzR17}}  \\
\multicolumn{2}{c|}{} &~~~PSNR~$\uparrow$~~~ & ~~~SSIM~$\uparrow$~~~ &~~~PSNR~$\uparrow$~~~ & ~~~SSIM~$\uparrow$~~~ \\
\toprule
\multirow{9}{*}{\textbf{CNN}~}&~CBDNet\cite{CBDNet} &30.78 & 0.801 &38.06 & 0.942\\

&~RIDNet\cite{RIDNet}  &38.71 & 0.951 &39.26 & 0.953 \\ 
&~AINDNet\cite{kim2020aindnet}  &39.08 & 0.954 &39.37 & 0.951 \\ 
&~VDN\cite{VDN}  &39.28 & 0.956 &39.38 & 0.952 \\ 

&~DeamNet\cite{ren2021adaptivedeamnet} &39.47 & 0.957 &39.63 & 0.953\\
&~CycleISP\cite{ZamirAKHK0020}  &39.52 & 0.957 &39.56 & 0.956 \\ 
&~MPRNet\cite{Zamir_2021_CVPR_mprnet} &39.71 & 0.958 &39.80 & 0.954\\
&~MIRNet\cite{ZamirAKHKYS20} &39.72 & 0.959 &\underline{39.88} & \textbf{0.956} \\ 
&~NAFNet\cite{chen2022simple_naf} &\textbf{40.30} & \textbf{0.961} &38.41 & 0.943 \\ 
\toprule
\multirow{1}{*}{\textbf{DM}}~&~\textbf{C2F-DFT (Ours)} ~~&\underline{39.84} & \underline{0.960} &\textbf{39.95} & \underline{0.955}  \\
\toprule
\end{tabular}
\end{center}
\end{table*}

\begin{figure*}[!t]
\centering
\includegraphics[scale=0.43]{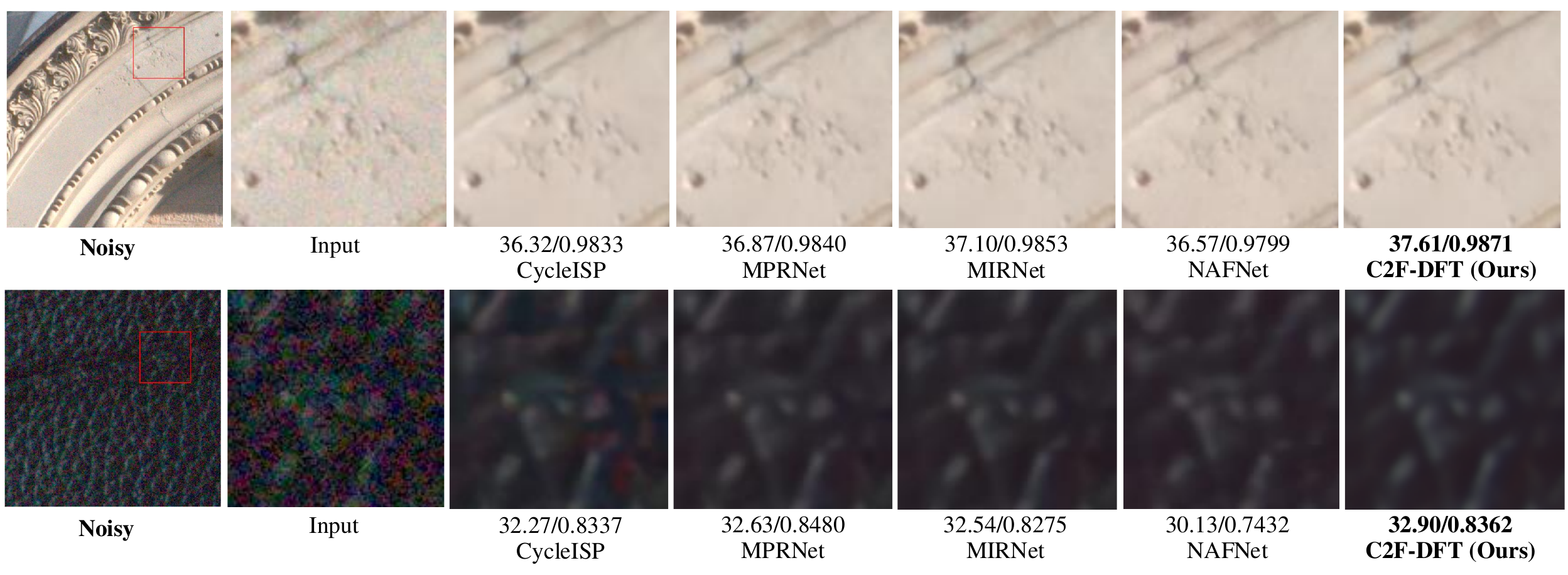}
\caption{
\underline{\textbf{Real image denoising on DND}}~\cite{PlotzR17}. C2F-DFT produces cleaner results with finer details.}
\label{fig:Real image denoising on DND}
\end{figure*}

\begin{figure*}[!t]
\centering
\includegraphics[scale=0.422]{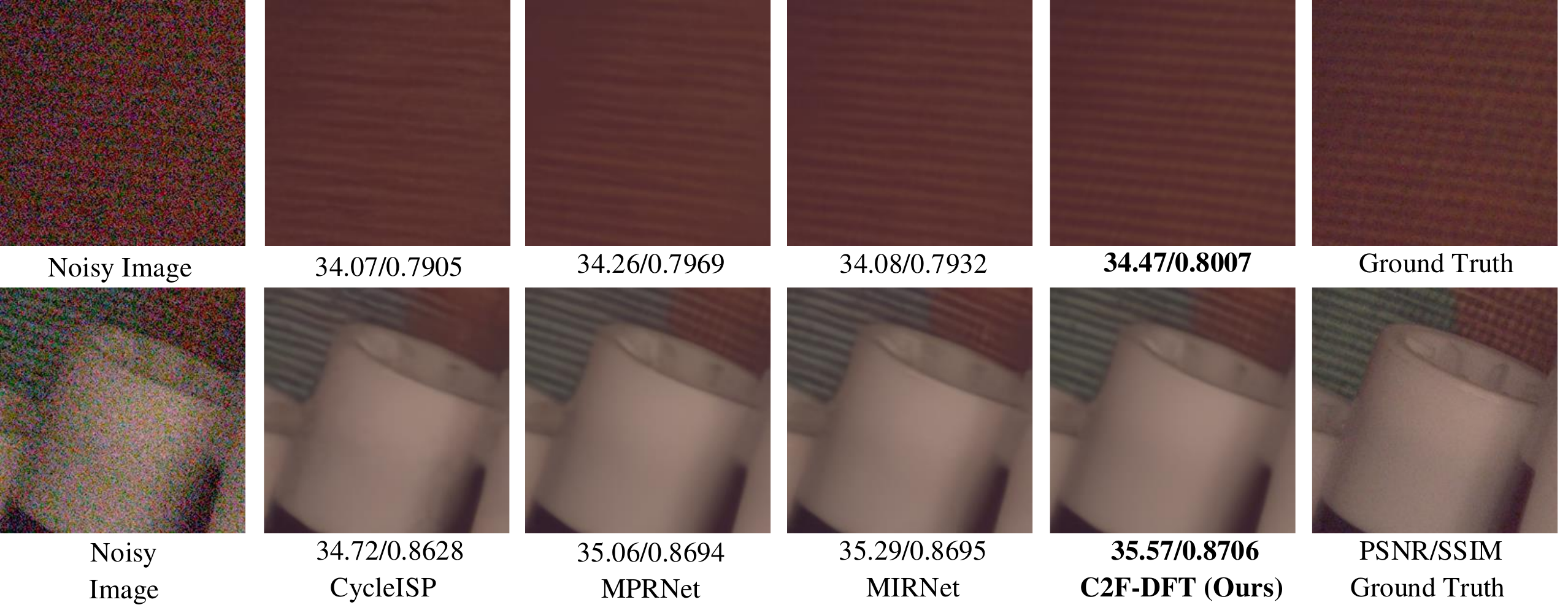}
\caption{
\underline{\textbf{Real image denoising on SIDD}}~\cite{sidd}. C2F-DFT achieves the best restoration quality.}
\label{fig:Real image denoising on SIDD}
\end{figure*}

\begin{table*}[t]
\begin{center}
\caption{\underline{\textbf{Comparison of LPIPS between IR-SDE and Our C2F-DFT on image deraining}}. 
$\uparrow$ and $\downarrow$ respectively denote higher and lower scores are better.
We use the publicly available codes to retrain IR-SDE on Rain13K dataset~\cite{Zamir2021Restormer}, then test on different test sets. The proposed C2F-DFT has the best perceptual performance for all rainy scenes.} \label{tab:LPIPS on deraining}
\setlength{\tabcolsep}{1.6mm}
\begin{tabular}{l|ccc|ccc|ccc|ccc}
\toprule
\multirow{2}{*}{\textbf{Method}}&\multicolumn{3}{c|}{\textbf{Test100}~\cite{ZhangSP20a}}  & \multicolumn{3}{c|}{\textbf{Rain100H}~\cite{derain_jorder_yang}} & \multicolumn{3}{c|}{\textbf{Rain100L}~\cite{derain_jorder_yang}}& \multicolumn{3}{c}{\textbf{Test2800}~\cite{0004P18}} \\
&PSNR~$\uparrow$&~SSIM~$\uparrow$~&LPIPS~$\downarrow$
&PSNR~$\uparrow$&~SSIM~$\uparrow$~&LPIPS~$\downarrow$
&PSNR~$\uparrow$&~SSIM~$\uparrow$~&LPIPS~$\downarrow$
&PSNR~$\uparrow$&~SSIM~$\uparrow$~&LPIPS~$\downarrow$
\\
\toprule
IR-SDE~\cite{irsde} 
&26.74&0.834&0.125
&20.79&0.699&0.267
&30.83&0.912&0.102
&30.42&0.891&0.065
\\
\textbf{C2F-DFT (Ours)} 
&\textbf{31.38}&\textbf{0.921}&\textbf{0.085}
&\textbf{31.62}&\textbf{0.909}&\textbf{0.125}
&\textbf{39.03}&\textbf{0.980}&\textbf{0.040}
&\textbf{34.03}&\textbf{0.944}&\textbf{0.047}
\\
\toprule
\end{tabular}
\end{center}
\end{table*}

\begin{table*}[t]
\begin{center}
\caption{\underline{\textbf{Comparison of LPIPS between IR-SDE and our C2F-DFT on image deblurring.}} Despite a slight inferiority on Gopro datasets in terms of LPIPS, our C2F-DFT outperforms IR-SDE and enjoys higher generalization on real datasets (RealBlur-R and RealBlur-J).} \label{tab:LPIPS on deblurring}
\setlength{\tabcolsep}{3.6mm}
\begin{tabular}{l|ccc|ccc|ccc}
\toprule
\multirow{2}{*}{\textbf{Method}}&\multicolumn{3}{c|}{\textbf{GoPro}~\cite{gopro2017} } & \multicolumn{3}{c|}{\textbf{RealBlur-R}~\cite{rim_2020_realblur} }& \multicolumn{3}{c}{\textbf{RealBlur-J}~\cite{rim_2020_realblur}}  \\
&PSNR~$\uparrow$&~SSIM~$\uparrow$&LPIPS~$\downarrow$
&PSNR~$\uparrow$&~SSIM~$\uparrow$&LPIPS~$\downarrow$
&PSNR~$\uparrow$&~SSIM~$\uparrow$&LPIPS~$\downarrow$
 \\
\toprule
IR-SDE~\cite{irsde}
&~30.70&0.901&\textbf{0.064}
&~33.96&0.918&0.114 
&~24.21&0.729&0.267
\\
\textbf{C2F-DFT (Ours)} 
&\textbf{31.96}&\textbf{0.928}&0.099
&\textbf{36.34}&\textbf{0.957}&\textbf{0.063}
&\textbf{28.90}&\textbf{0.876}&\textbf{0.153}
\\
\toprule
\end{tabular}
\end{center}
\end{table*}

\subsection{Main Results}

\subsubsection{Image Deraining}
We conduct deraining experiments on the Rain13K dataset~\cite{Zamir2021Restormer}, which contains over 13,712 clean-rainy image pairs. For evaluation, the trained model
is tested on the full-size test images of four synthetic datasets: Rain100H~\cite{derain_jorder_yang}, Rain100L~\cite{derain_jorder_yang}, Test100~\cite{ZhangSP20a}, and Test2800~\cite{0004P18}. Similar to existing methods~\cite{mspfn2020,Zamir_2021_CVPR_mprnet,purohit2021spatially_spair,Zamir2021Restormer}, we report PSNR/SSIM scores using Y channel in YCbCr color.
Tab.~\ref{tab:Image deraining.} summarises quantitative comparisons, where our method consistently outperforms CNN-based methods, e.g., \cite{purohit2021spatially_spair}, and is competitive with Transformer-based methods~\cite{Zamir2021Restormer}.
Compared with the diffusion model-based method IR-SDE~\cite{irsde}, our method achieves a significant gain, increasing $6.81$dB PSNR when averaging these benchmarks.
Fig.~\ref{fig:Image deraining on Rain100L}   presents two visual examples on Rain100L~\cite{derain_jorder_yang} and Test100~\cite{ZhangSP20a}, where our C2F-DFT is able to recover a much clearer result. 

\subsubsection{Image Deblurring}
Following~\cite{Zamir2021Restormer}, we train our method on GoPro~\cite{gopro2017} dataset, which contains 2103 clean-blur image pairs, and then directly apply to real benchmarks RealBlur-R and RealBlur-J~\cite{rim_2020_realblur} datasets.
Tab.~\ref{tab:Image deblurring} shows that our method is slightly inferior to existing methods on the GoPro but shows a strong generalization to RealBlur datasets.
It is worth noticing that our method achieves the SOTA on RealBlur-R and is comparable with Restormer~\cite{Zamir2021Restormer} on RealBlur-J.
Especially, our method significantly outperforms the diffusion model-based method IR-SDE~\cite{irsde} trained on GoPro by $4.69$dB PSNR on RealBlur-J, demonstrating our method is a better diffusion deblurring model.
Fig.~\ref{fig:Image deblurring on RealBlur-R} shows that our method is able to generate a sharper result with finer structures on RealBlur-R and RealBlur-J test sets.
These qualitative and quantitative results clearly show that our method is a strong deblurring diffusion model, especially on unseen scenes.

\subsubsection{Real Image Denoising}
%
Following~\cite{Zamir2021Restormer}, our method is trained on SIDD~\cite{sidd} dataset which contains 320 high-resolution image pairs. The training datasets are randomly cropped into 30,608 image patches with $512 \times 512$ size. We evaluate 1,280 patches with size $256 \times 256$ from the SIDD~\cite{sidd} validation set and 1,000 patches with size $512 \times 512$ from the DND~\cite{PlotzR17} benchmark dataset. Note that since the Ground Truth of the DND dataset is not publicly available, the results are online evaluated at \url{https://noise.visinf.tu-darmstadt.de/}.
Tab.~\ref{tab:Real image denoising} shows that our method is comparable with recent state-of-the-art NAFNet~\cite{chen2022simple_naf} on SIDD, but outperforms it over $1.54$dB on DND in terms of PSNR, which adequately illustrates our diffusion restoration method has a better generalization to unseen scenes, the same conclusion on image deblurring.
We further provide the visualization results on DND in Fig.~\ref{fig:Real image denoising on DND}, where our method has favorable denoising capability while preserving sharper structures by observing the locally enlarged areas. Fig.~\ref{fig:Real image denoising on SIDD} shows visual comparisons and corresponding PSNR/SSIM scores on the SIDD~\cite{sidd} dataset, in which our C2F-DFT achieves the best restoration quality.

\subsubsection{Perceptual Measurement Comparisons between IR-SDE and Our C2F-DFT}
%
Following IR-SDE~\cite{irsde}, we introduce a perceptual measurement LPIPS~\cite{ZhangIESW18} to measure the perceptual image restoration quality. Tab.~\ref{tab:LPIPS on deraining} and Tab.~\ref{tab:LPIPS on deblurring} respectively show the results on image deraining and image deblurring tasks.
The results demonstrate that our C2F-DFT can consistently outperform the IR-SDE~\cite{irsde} on perceptual measurement, suggesting the effectiveness of perceptual measurement of our C2F-DFT.

\begin{table}[!t]
\begin{center}
\caption{\underline{\textbf{Sampling steps $S$}} for image restoration.}
\label{tab:Effect on different sampling steps tested on the entire image.}
\setlength{\tabcolsep}{1.15mm}
\begin{tabular}{l|cc|c|c}
\toprule
\textbf{$S$}~~~&~~~~~PSNR~$\uparrow$~~~~&~~~~~SSIM~$\uparrow$~~~~&~~~~~LPIPS~$\downarrow$~~~~~ &~~~~~Time (s)~~~~~\\
\toprule
2	&10.99 & 0.019	&1.4166 &31\\
3	&25.61 & 0.878	&0.1588 &58\\
\textbf{4} &\textbf{30.93} & \textbf{0.900}	& 0.1313 &102\\
5	&30.44 & 0.894	&0.1279 &129\\
10 &30.00 & 0.884	&0.1231 &237\\
25 &29.30 & 0.868	&0.1138 &682\\
50 &28.98 & 0.857	&0.1092 &818\\
\toprule
\end{tabular}
\end{center}
\end{table}

\begin{table}[!t]
\begin{center}
\caption{\underline{\textbf{Fixed patch vs. patch-cycle}} on diffusion training.
For the fixed patch, we use the fixed $64 \times 64$ patch to train the diffusion model, while we use $\{32 \times 32, 64 \times 64, 128 \times 128 \}$ patch to cyclically conduct the training process.
}\label{tab:Effect of the fixed patch and patch-cycle diffusion training.}
\setlength{\tabcolsep}{1.5mm}
\begin{tabular}{l|ccc}
\toprule
\textbf{Training Patch} ~~& ~~PSNR~$\uparrow$~~&~SSIM~$\uparrow$~~&~~Training Time (h)~ \\
\toprule
(a) Fixed Patch 	&30.52&0.896&84.2 \\
\textbf{(b) Patch-Cycle (Ours)}  &\textbf{30.93}&\textbf{0.900}&90.3\\
\toprule
\end{tabular}
\end{center}
\end{table}
\begin{figure}[!t]
\centering
\includegraphics[scale=0.28]{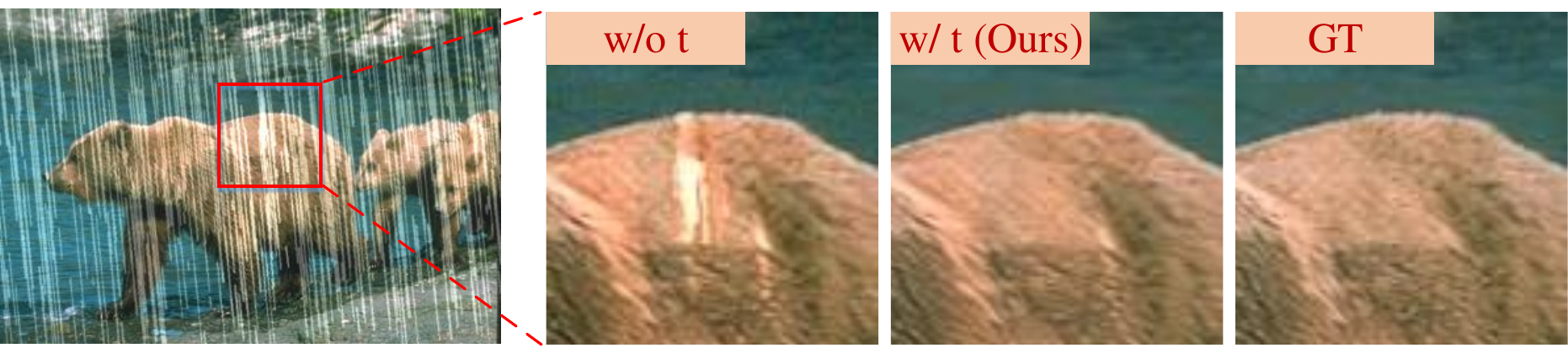}
\caption{
\underline{\textbf{Without vs. with $t$ embedding}} visual results. Disabling $t$ embedding results in handing down extensive noises, while embedding $t$ into Transformers can better remove noise and restore clearer images.}
\label{fig:With vs. without t embedding visual results}
\end{figure}

\begin{table*}[t]
\begin{center}
\caption{\underline{\textbf{Coarse-to-fine training pipeline}} analysis on the $3$ rest deraining test sets.
Our fine training strategy, which is optimized by constraining sampled restoration results instead of noises, consistently improves the restoration quality of coarse training that optimizes the diffusion models by constraining the noises.
} 
\label{tab:Coarse-to-fine training pipeline analysis}
\setlength{\tabcolsep}{3.35mm}
\begin{tabular}{l|cc|cc|cc|cc}
\toprule
\multirow{2}{*}{\textbf{Training Iteration}} & \multicolumn{2}{c|}{\textbf{Test100}~\cite{ZhangSP20a}}  & \multicolumn{2}{c|}{\textbf{Rain100H}~\cite{derain_jorder_yang}} & \multicolumn{2}{c|}{\textbf{Rain100L}~\cite{derain_jorder_yang}} & \multicolumn{2}{c}{\textbf{Test2800}~\cite{0004P18}} \\
&PSNR~$\uparrow$&SSIM~$\uparrow$ 
&PSNR~$\uparrow$&SSIM~$\uparrow$ 
&PSNR~$\uparrow$&SSIM~$\uparrow$
&PSNR~$\uparrow$&SSIM~$\uparrow$
\\
\toprule
(a) Coarse 270K 
&30.94&0.914	
&30.93&0.900	
&38.47&0.975	
&33.73&0.941 \\
(b) Coarse 360K 
&31.11&0.915	
&30.97&0.900	
&38.44&0.975	
&33.74&0.941 \\
\textbf{(c) Coarse 270K + Fine 90K (Ours)} &\textbf{31.38}&\textbf{0.921}	&\textbf{31.62}&\textbf{0.909}	
&\textbf{39.03}&\textbf{0.980}	
&\textbf{34.03}&\textbf{0.944} 
\\
\toprule
\end{tabular}
\end{center}
\end{table*}

\begin{figure*}[t]
\centering
\includegraphics[scale=0.35]{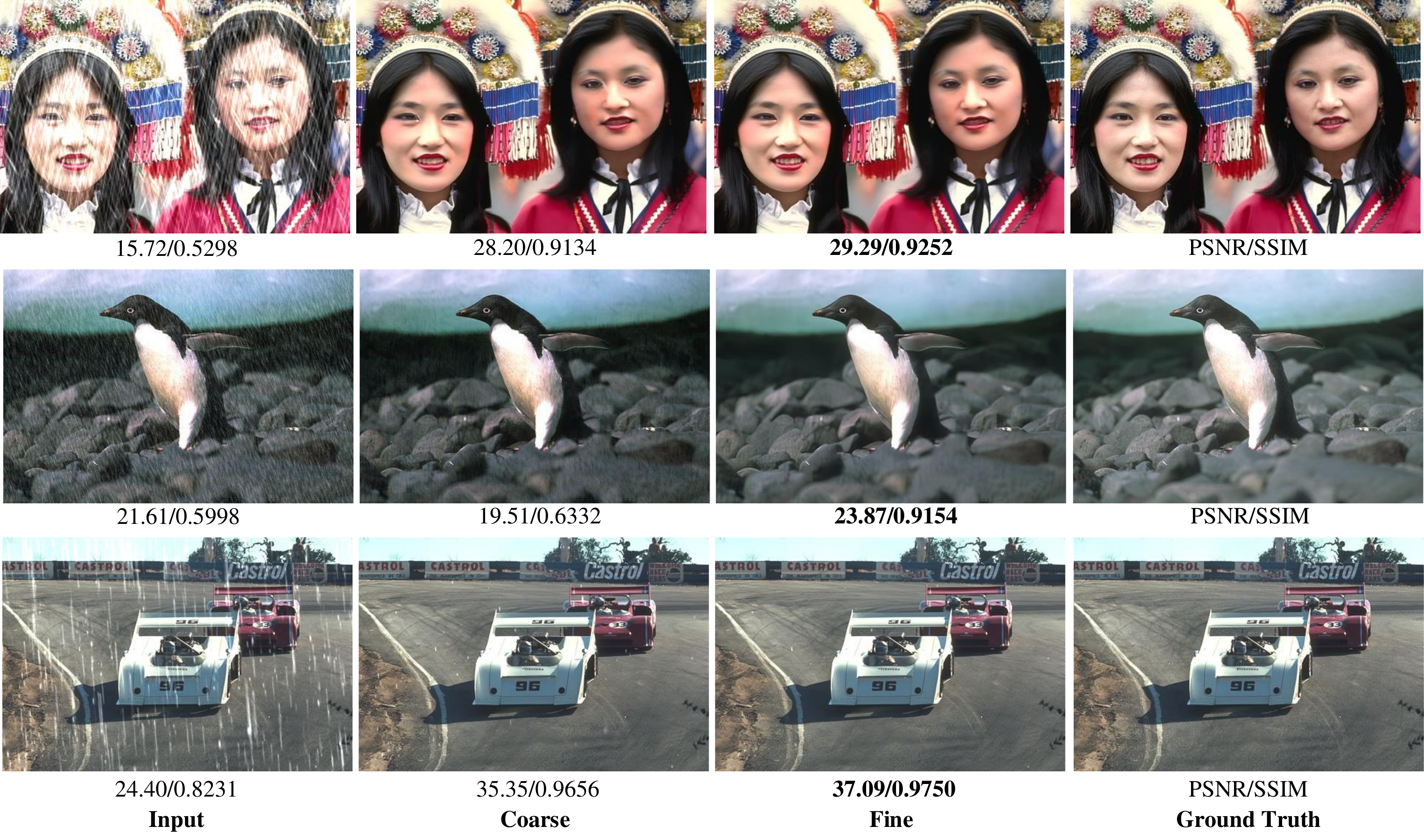}
\caption{\underline{\textbf{Coarse training vs. fine training visual results}} on image draining.
Our fine training strategy can effectively improve the restoration quality compared with the results of coarse training.
}
\label{fig:Coarse vs. Fine training visual results on image deraining.}
\end{figure*}

\begin{figure*}[!t]
\centering
\includegraphics[scale=0.35]{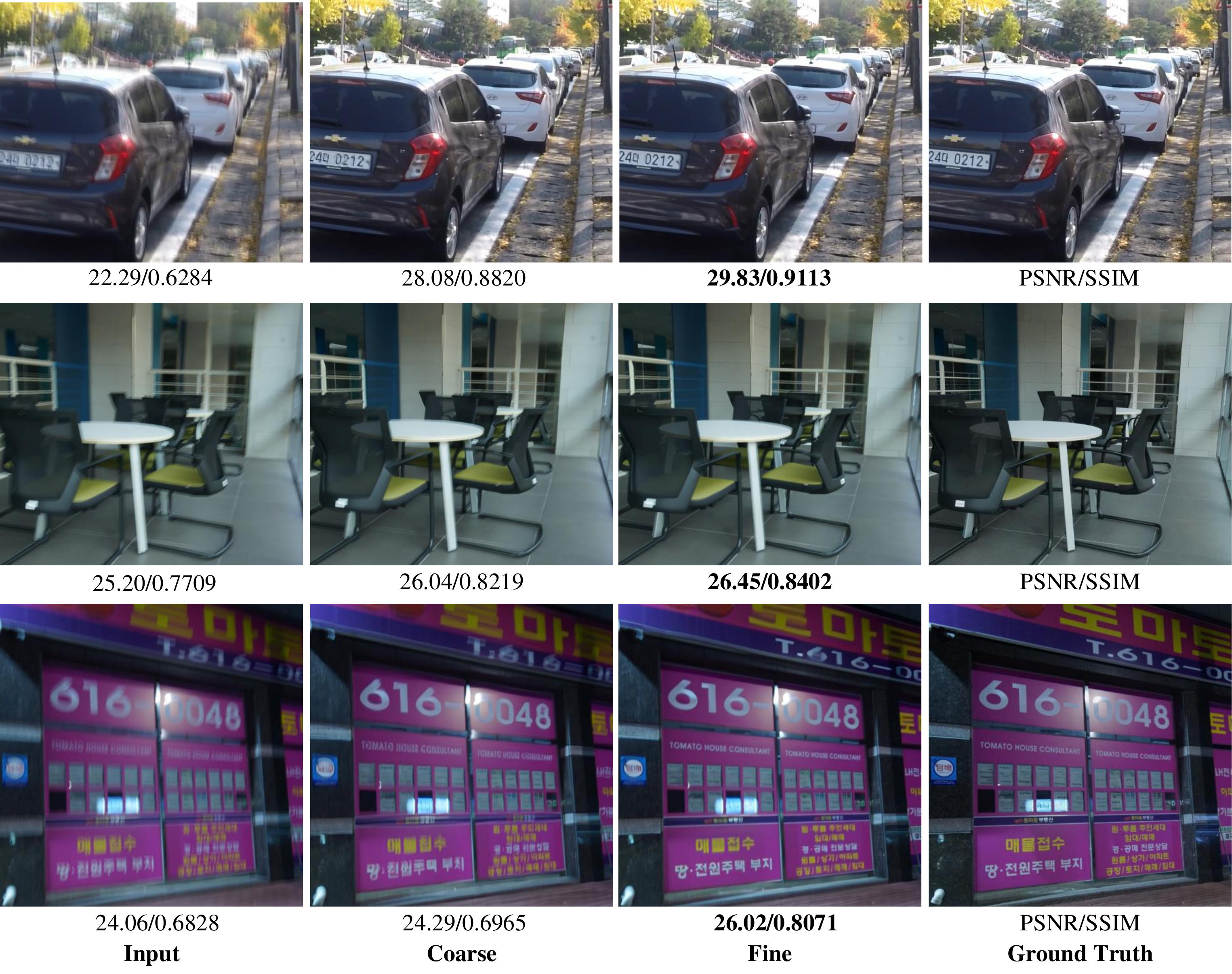}
\caption{\underline{\textbf{Coarse training vs. fine training visual results}} on image deblurring.
Our fine training strategy can effectively improve the restoration quality compared with the results of coarse training.}
\label{fig:Coarse vs. Fine training visual results on image deblurring.}
\end{figure*}

\subsection{Ablation Study}
We conduct the ablation study to verify the effectiveness of the proposed components on the Rain100H test set.

\subsubsection{Effect on Sampling Steps}
Given that diffusion models involve sampling images iteratively until achieving satisfactory results, it becomes essential to analyze the impact of different sampling steps (denoted as S) on restoration quality. 
In Tab.~\ref{tab:Effect on different sampling steps tested on the entire image.}, we present the distortion metrics PSNR/SSIM and the perception measurement LPIPS, along with the corresponding sampling time. 
Notably, fewer sampling steps lead to higher distortion metrics (PSNR and SSIM), whereas more sampling steps yield improved perception results (LPIPS) at the cost of increased processing time. 
This observation aligns with the findings in~\cite{WhangDTSDM22}. 
By carefully balancing these metrics and considering sampling time, we opt to utilize $4$ sampling steps in constructing the fine training pipeline.

\subsubsection{Effect on Patch-Cycle}
We employ the patch-cycle training strategy to train our model, which raises questions about its effectiveness and impact on restoration quality. 
To address these concerns, we conduct separate training experiments using patch-cycle and fixed-patch approaches during the coarse training phase. 
The results in Tab.~\ref{tab:Effect of the fixed patch and patch-cycle diffusion training.} demonstrate that our method outperforms fixed patch-based training while requiring a similar amount of training time.

\subsubsection{Effect on Time Embedding}
%
As we embed time step $t$ into both DFSA and DFN in Transformers, one may wonder about its effect on restoration.
Tab.~\ref{tab:With t embedding vs. without t embedding.} shows that embedding $t$ to Transformers can significantly improve the restoration quality by $4.08$dB PSNR.
The results reveal that time embedding is critical for diffusion-based image restoration methods.
Fig.~\ref{fig:With vs. without t embedding visual results} shows that disabling $t$ embedding hands down extensive noises, while embedding $t$ to Transformers can significantly improve the visual restoration quality.

\begin{table}[h]
\begin{center}
\caption{\underline{\textbf{Without vs. with $t$ embedding}} effect on our DFT.}\label{tab:With t embedding vs. without t embedding.}
\setlength{\tabcolsep}{2mm}
\begin{tabular}{l|ccc}
\toprule
\textbf{Experiment} &~~~~~~~PSNR~$\uparrow$~~~~~~~&~~~~~~SSIM~$\uparrow$~~~~~~~\\
\toprule
(a) w/o $t$ embedding	&26.85&0.840 \\
\textbf{(b) w/~$t$ embedding (Ours)}  &\textbf{30.93}&\textbf{0.900}\\
\toprule
\end{tabular}
\end{center}
\end{table}

\subsubsection{Effect on Coarse-to-Fine Training Strategy}
%
To better illustrate the effectiveness of our coarse-to-fine training pipeline, we further analyze the effect of the proposed coarse-to-fine training pipeline on four rainy test sets in Tab.~\ref{tab:Coarse-to-fine training pipeline analysis}.
One can obviously observe that more iterations on coarse training only bring minor improvement (0.04dB on Rain100H and 0.01dB on Test2800 in Tab.~\ref{tab:Coarse-to-fine training pipeline analysis}(a) vs. (b)).
However, adding fine training on the completed coarse training can significantly improve the restoration results, compared with the only coarse training versions (Tab.~\ref{tab:Coarse-to-fine training pipeline analysis}(c) vs. (a)-(b)).
Fig.~\ref{fig:Coarse vs. Fine training visual results on image deraining.} and Fig.~\ref{fig:Coarse vs. Fine training visual results on image deblurring.} show some examples of the comparisons between coarse and fine training on image deraining and image deblurring, 
where our fine training can effectively improve image restoration quality, while the coarse training always hands down extensive noises or artifacts.
This may be because using the fine training that constrains the sampled restored results with ground truth instead of noise to optimize the diffusion model can help the model learn more complex degraded information, which would facilitate the model to better restore clean images.




%

\section{Conclusion}
In this paper, we have proposed a diffusion Transformer with a new coarse-to-fine training scheme for image restoration. 
Observing that restoration quality may be affected due to inaccurate noise estimation in diffusion models, we have proposed the coarse-to-fine training scheme to improve the restoration quality by constraining the sampled restoration results instead of noises, enabling to facilitate better restoration in the fine training stage. 
Extensive experiments show that our C2F-DFT achieves competitive performance compared with state-of-the-art methods on $3$ image restoration tasks, including image deraining, image deblurring, and real image denoising.


{
\bibliographystyle{IEEEtran}
\bibliography{paper.bib}
}

\end{document}